\documentclass{article}

\usepackage{arxiv}

\usepackage[utf8]{inputenc} % allow utf-8 input
\usepackage[T1]{fontenc}    % use 8-bit T1 fonts
\usepackage{hyperref}       % hyperlinks
\usepackage{url}            % simple URL typesetting
\usepackage{booktabs}       % professional-quality tables
\usepackage{amsfonts}       % blackboard math symbols
\usepackage{nicefrac}       % compact symbols for 1/2, etc.
\usepackage{microtype}      % microtypography
\usepackage{graphicx}
\usepackage[numbers]{natbib}
\usepackage{doi}

\usepackage{soul}
\usepackage{pifont}
\newcommand{\cmark}{\ding{51}}%
\newcommand{\xmark}{\ding{55}}%

\newcommand{\vars}[1] {\ensuremath{\mathbf {#1}}}
\newcommand{\var}[1] {\ensuremath{#1}}
\newcommand{\graph}[1] {\ensuremath{\mathcal {#1}}}

\usepackage{multirow}

\usepackage{subfig}

\title{Towards Automated Causal Discovery: a case study on 5G telecommunication data}

\author{ 
    Konstantina Biza \\ 
	University of Crete\\
    FORTH\\
	Heraklion, Greece \\
	\And
	Antonios Ntroumpogiannis \\
	University of Crete\\
    FORTH\\
	Heraklion, Greece \\
	\And
    Sofia Triantafillou \\ 
	University of Crete\\
    FORTH\\
	Heraklion, Greece \\
	\And
	Ioannis Tsamardinos \\
	University of Crete\\
    FORTH\\
	Heraklion, Greece \\
}

\begin{document}
\maketitle

\begin{abstract}
We introduce the concept of Automated Causal Discovery (AutoCD), defined as any system that aims to fully automate the application of causal discovery and causal reasoning methods. AutoCD's goal is to deliver all causal information that an expert human analyst would and answer a user's causal queries. We describe the architecture of such a platform, and illustrate its performance on synthetic data sets. As a case study, we apply it on temporal telecommunication data. The system is general and can be applied to a plethora of causal discovery problems. 
\end{abstract}


\section{Introduction}
Causal Discovery is a field of machine learning and statistics aiming to induce causal knowledge from data \cite{Pearl2009, Spirtes2000}. There is a large corpus of algorithms and methodologies in the field, spanning tasks like learning  causal models, estimating causal effects, and determining  optimal interventions. While there are several public libraries of algorithms for these tasks, combining the algorithms and applying them to any given problem is a challenging endeavor that requires extensive knowledge of the methods and a deep understanding of the theory to interpret results.

In this paper, we introduce the concept of Automated Causal Discovery ({\bf AutoCD}) (not to be confused with Automated Causal Inference \cite{CausalTune, Nguyen2023}; see Section \ref{sec:relatedwork}), defined as the effort to fully automate the application of causal discovery and causal reasoning. AutoCD's goals should be to deliver not just the optimal causal model that fits the data, but all information, answers to queries, visualizations, interpretations, and explanations that a human expert analyst would. AutoCD is meant to make an esoteric field and its methodologies accessible to non-experts. These are prone to errors and produce results whose interpretation requires a deep understanding of the causal modeling theory. Just like in the term AutoML (automated machine learning) for predictive modeling, the use of the term ``Auto'' implies that, among others, an optimization of the (causal) machine learning pipeline is taking place. 

Working towards this goal, we propose an AutoCD architecture, shown in Figure \ref{fig:architecture}.
This architecture is capable of dealing with high-dimensional and mixed-type data, that are either temporal or cross-sectional. It optimizes the data representation, the causal discovery algorithm, and its hyper-parameters to learn the causal model that best fits the data, out of a palette of state-of-the-art algorithms. Finally, it employs the Markov Equivalence class of the model to answer user-defined causal queries and visualize results.  

Our contribution in this work is twofold. First, we implement and introduce AutoCD, a library for various causal learning and reasoning tasks targeting non-experts users. Secondly, we apply AutoCD on a real-world telecommunication problem. 

The rest of the document is organized as follows. In section \ref{sec:background}, we briefly present the basic concepts of causal discovery to enable an autonomous reading of the paper.  In section \ref{sec:relatedwork}, we refer to the related work and discuss the similarities and the differences with AutoCD. In section \ref{sec:autocd}, we describe the AutoCD architecture. In section \ref{sec:problem} we describe the telecommunication case-study and the main challenges of applying causal learning. In addition, we  report the results of the AutoCD's application on the real data. In sections \ref{sec:exp_resim} and \ref{sec:exp_synthetic}, we provide an experimental benchmarking of our framework on resimulated data from the telecommunication problem and synthetic data, respectively. Finally, in section \ref{sec:future}, we discuss the limitations and future work of AutoCD.

\section{Background}
\label{sec:background}

Two types of causal graphs are commonly used to represent the causal relationships between the variables, the Directed Acyclic Graph (DAG) and the Marginal Ancestral Graph (MAG). When the causal DAG is annotated with conditional probability densities, it becomes a quantitative causal model, namely a Bayesian Network (BN). In the rest of the paper, we'll be using the terms DAG and BN interchangeably, when the densities are not important. A DAG contains only directed edges, which represent direct causality, i.e. if $ X \longrightarrow Y$  then $X$ is a direct cause of $Y$ (direct in the context of the observed variables). A MAG contains both directed and bi-directed edges. MAGs admit and represent latent confounding; they used to be employed when latent confounding is possible. We refer to such cases as causally insufficient systems. The interpretation of edges in a MAG is not as straightforward as in DAG models. In a MAG, an edge $X \longrightarrow Y$  denotes that X causes Y, but the causal relation may not be direct and it may be confounded by latent variables (see \cite{borbudak_magedges,Triantafillou2014ConstraintbasedCD, richardson2002} for a deeper look into the interpretation of MAG edge semantics). An edge $X \longleftrightarrow Y$ means that  $X$ does not cause  $Y$ and $Y$ does not cause $X$, therefore the two share a latent common cause.

Both DAGs and MAGs capture the independence relationships between the observed variables \cite{Pearl2009, Spirtes2000, Neapolitan2003, richardson2002, Zhang2008CausalRW, borbudak_magedges}.  
Different graphs over the same set of variables can entail the same conditional independencies. These graphs are called {\bf Markov Equivalent} graphs. {\em Markov Equivalent graphs represent different causal theories that equally well fit the data (their conditional independences to be specific).} 
A Partially Directed Acyclic Graph (PDAG) represents a class of Markov Equivalent DAGs. Similarly, a Partial Ancestral Graph (PAG) represents a class of Markov Equivalent MAGs. Both PDAGs and PAGs contain only the edges (ignoring direction) shared by all members of the equivalence class. The end-points of the edges in MAGs can be either arrows or circles. An arrow denotes that the edge has the specific direction shared by all members of the class, while a circle indicates disagreement between members: a circle could be an arrow or a tail. For example, the edge $\var A \circ \rightarrow \var B$ in a PAG denotes that there exist members of the class with the edge $\var A \rightarrow \var B$ as well as $\var A \leftrightarrow \var B$. The interpretation of the edges is that $\var A$ is causally affecting $\var B$ (possibly indirectly and possibly the relation is also confounded by a latent variable), or, neither variable is causing each other and there exist a latent confounding variable \cite{Richardson2002AncestralGM, Zhang2008CausalRW}.

In a causal graph,  the Markov boundary (\vars {Mb}) of a node $X$ consists of its adjacent nodes and the nodes that are reachable through a collider path \cite{Borboudakis2019ForwardBackwardSW, Pellet2008FindingLC}. The $\vars {Mb} (X)$ has the following important properties: {\em it is the minimal set that renders \var X conditionally independent of any other node; it is the solution to the feature selection problem when trying to predict \var X under certain broad conditions \cite{tsamardinos03a}.} 
It is shown that in distributions faithful to a BN (see \cite{Spirtes2000} for a definition of faithfulness), the Markov boundary is unique. Hence, under faithfulness, different equivalent MAGs/DAGs may disagree in the direction of causal edges, but agree on all Markov boundaries (all \vars{Mb} are invariant). 

A wide range of causal discovery algorithms learn a causal structure from observational data. Algorithms that assume causal sufficiency estimate a single DAG or its corresponding PDAG, and algorithms that allow causal insufficiency induce a MAG or its corresponding PAG  \cite{Zhang2008CausalRW}. 
Causal discovery algorithms are mainly grouped in three categories. Constraint-based methods identify conditional independencies using statistical tests, and try to reverse-engineer the graphs that represent these independencies \cite{Spirtes2000,Ramsey2006AdjacencyFaithfulnessAC,Colombo2014OrderindependentCC, Raghu2018ComparisonOS, Colombo2012LearningHD}. Score-based methods try to identify the graphical model that best fits the data, using some scoring function\cite{Cooper1992, Chickering2002OptimalSI, Ramsey2016AMV}. Hybrid algorithms combine the constraint-based approaches with the scoring techniques \cite{ogarrio16, Tsamardinos2006TheMH, Tsirlis2018OnSM}. There are also causal discovery algorithms that are not included in the above categories  \cite{Shimizu2006ALN, Zheng2018DAGsWN, Kalainathan2018StructuralAM}.  
Extensions of the above have been also proposed for temporal data \cite{Runge2019DetectingAQ,Runge2020DiscoveringCA,Gerhardus2020HighrecallCD,Malinsky2018CausalSL,hyvarinen10a,pamfil20_dynotears,peters_timino}. For causal discovery on temporal data, most of the corresponding algorithms assume stationarity, that implies that the causal relationships (causal structure) remain the same across the time lags \cite{Runge2018CausalNR}.

\section{Related Work}
\label{sec:relatedwork}

AutoCD addresses the automation of causal discovery and causal reasoning. There are various libraries of causal analysis-related algorithms (such as Tetrad \footnote{https://github.com/cmu-phil/tetrad} and  Tigramite \footnote{https://github.com/jakobrunge/tigramite}), but they do not constitute automated analysis systems. The three works most relevant to AutoCD are  PyWhy \footnote{http://pywhy.org},  OpportunityFinder \cite{Nguyen2023} and the CausalMGM \cite{Ge2020CausalMGMAI}.

PyWhy is a system for causal learning and inference with several libraries including causal-learn \cite{causallearn} for causal discovery, DoWhy \cite{dowhy, dowhy_gcm} for causal effect estimation and what-if questions, EconML \cite{econml} for conditional causal effect estimation and CausalTune \cite{CausalTune} for selecting the optimal approach for causal effect estimation.
AutoCD and PyWhy share a long list of causal discovery algorithms and a set of general ideas, such as the use of a hyper-parameter tuning method, methods to test if the obtained result is valid, and interpretability tools. 
However, at this moment, the two works focus on different aspects of analysis: AutoCD focuses on causal discovery and PyWhy on causal inference. As a result, they significantly differ in practice.

OpportunityFinder \cite{Nguyen2023} is a causal inference framework, which aims to automate the estimation of the causal effect, inspired by the AutoML community. Given treatment and observational datasets, it pre-processes the inputs, selects and runs the best algorithm for causal effect estimation, and validates the output.  
Both AutoCD and OpportunityFinder are proposed for non-experts users and have common modules, such as the selection of the most suitable algorithm. However, as with PyWhy, the goal of OpportunityFinder is to return the causal effect of a given treatment, while AutoCD currently focuses on finding the optimal causal graph and reasoning with it.

CausalMGM \cite{Ge2020CausalMGMAI} is a web-based tool for causal discovery that performs feature selection, causal discovery, and visualization. These three modules appear also in AutoCD, however, we point out the following differences: (a) CausalMGM applies a single feature selection algorithm, namely Pref-Div \cite{Ge2020EfficientPA}, while AutoCD has an automated pipeline to select the optimal feature selection method among an extendable set of algorithms. (b) AutoCD uses causal tuning \cite{Biza2022OutofSampleTF} to automate causal discovery, selecting from a variety of causal discovery algorithms and hyper-parameters, while CausalMGM uses the StEPS \cite{Sedgewick2016LearningMG} tuning procedure to select optimal hyper-parameters of a single algorithm, the MGM method. (c) AutoCD allows for causally insufficient systems, while CausalMGM assumes no latent confounding, and (d) in AutoCD we include edge confidence estimations using the estimated causal structure. On the other hand,  CausalMGM has an interactive web platform that allows a non-expert user to apply causal discovery and visualize the results without any prior knowledge. A web-based tool is one of the future goals for AutoCD.

\section {AutoCD architecture}
\label{sec:autocd} 

\begin{figure}
  \centering
  \includegraphics[width=0.6\linewidth]{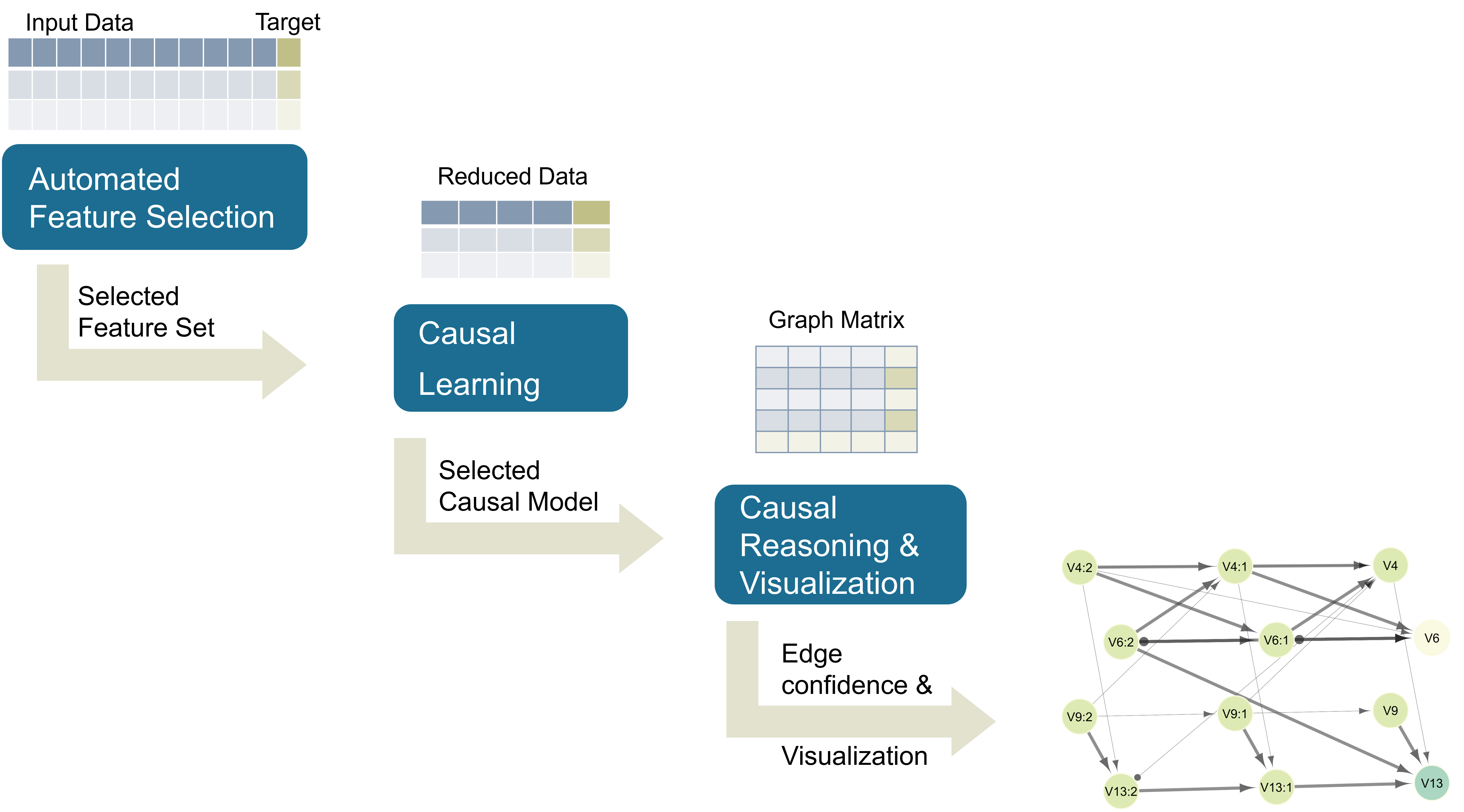}
  \caption{The proposed architecture for Automated Causal Discovery.}
  \label{fig:architecture}
\end{figure}

Figure \ref{fig:architecture} shows our proposed methodology for AutoCD. The architecture consists of three main modules:  Automated Feature Selection (AFS),  Causal Learning (CL), and  Causal Reasoning and Visualization (CRV). In the first step, AFS reduces the dimensionality of the problem, by only filtering in variables that are necessary for optimizing an optional set of user-defined desired outcomes. In the second step, CL learns a causal model over the features selected in the first step, by optimizing the causal analysis pipeline.  In the third step,  CRV visualizes, summarizes, explains, and interprets this model as a response to a set of user-defined queries. We now explain these modules in turn.

\subsection{Automated Feature Selection}

The purpose of the AFS module is to reduce the number of features required to be considered for subsequent causal modeling and inference. Specifically, in high-dimensional datasets, the user can optionally specify an outcome of interest; AFS then seeks to identify its \vars {Mb}. The \vars {Mb} returned by AFS will be used in the second module that learns a causal model. This guarantees that direct parents of the outcome will not be filtered out.
However, we note that filtering out features and learning the causal model of the marginal distribution, may result in missed opportunities for learning the directionality of causal edges. 

The input to AFS is (a) a dataset and (b) an outcome of interest. The output of AFS are (i) estimates of the Markov boundary of the outcome (feature selection results), (ii) a predictive model for the outcome using its \vars {Mb}, and (iii) an out-of-sample estimate of the predictive performance of the model with respect to  the outcome.

AFS is implemented using an automated machine learning platform for predictive modeling, equipped with feature selection. The automated machine-learning approach implies the machine-learning pipeline (called a  {\bf configuration}, hereafter) is tuned and optimized for the data and outcome at hand. This is an important factor in achieving high performance in terms of selecting the best approximations of the \vars {Mb}. AFS's configurations comprise of two steps: feature selection, and modeling. A Combined Algorithm Selection and Hyperparameter (CASH) \cite{Thornton2012AutoWEKACS} optimization takes place, where a search is performed in the space of possible configurations.

We now explain in more detail the current implementation of AFS. We follow the design principles of JADBio (Just Add Data Bio, see Figure 6 in \cite{Tsamardinos2022JustAD}), a commercial AutoML tool that includes feature selection, performs CASH optimization and reports an unbiased estimation of performance. First, a subsystem (called Algorithm and Hyperparameter Space selection, or AHPS) selects the applicable algorithms to run on the data and the ranges of their hyperparameter. It also selects the protocol to estimate the performance of each configuration, e.g., to run a repeated, 5-fold cross-validation for small samples or imbalanced data, or to run a single hold-out for larger datasets. In JADBio, AHPS makes these decisions based on a rule-based system, augmented with rules from meta-level learning. In AFS, these decisions are hard-coded for simplicity: AFS configuration space is always the same for all datasets. For feature selection, AFS employs the FBED, and SES algorithms \cite{FBED, Lagani2016FeatureSW} from the MXM R Package \cite{MxMPackage}. These algorithms scale up to tens of thousands of features, apply on any several types of outcomes (nominal, ordinal, continuous, time-to-event), and return multiple statistically equivalent solutions, i.e., multiple possible \vars{Mb} whose predictive performance is statistically indistinguishable. The algorithms also have well-defined causal properties. Specifically, FBED is guaranteed to return the \vars{Mb} asymptotically, in faithful distributions, assuming perfect tests of conditional independence, while SES is guaranteed to return the causal neighbors (direct causes and direct effects) of the outcome, under the same conditions. The estimation protocol in AFS is always 5-fold Cross-Validation. Regarding modeling algorithms, the current implementation of AFS optimize over hyper-parameters of the Random Forests algorithm.

Subsequently, the Configurator Generator (CG) generates the configurations to try within the configuration space defined by the AHPS in the previous step. In the current implementation, the CG performs a simple grid search in the Cartesian product of all hyper-parameters of each algorithm  (JADBio uses more sophisticated CASH methodologies).  Next, the Configuration Evaluator (CE) uses the selected estimation protocol to determine the predictive performance of a configuration. The AUROC, the average AUROC of each class-vs-rest, and the coefficient of determination $R^2$ are the performance metrics of choice to optimize, for binary, multi-class, and continuous outcomes, respectively.

Applying the winning configuration on all available data returns the final selection of features (\vars {Mb}) and produces the final model, expected to be the best performing on average.

\subsection{Causal Learning}

The CL module is analogous in design to the AFS, in the sense that it searches in the space of configurations to identify the one with the highest performance. The notable difference, of course, is that the configuration space includes configurations that produce causal models and not predictive models. As in AFS, an AHPS system decides which algorithms to apply to a given problem and the ranges of their hyper-parameters. In CL these are hard-coded for simplicity. Table \ref{table:causalconfigs}  presents the causal algorithms, currently available in the system, with the following information: whether they admit the presence of latent confounders  (\textit{Latent}), and whether they have been proposed for temporal data and enforce stationarity constraints (\textit{Temp.}). Column named \textit{Tests/Scores} shows the type of supported conditional independence tests and scoring functions employed internally by the algorithms. These tests and scores indicate the type of assumptions made by the algorithm (e.g., linearity) and the type of data to which they apply. The data type is explicitly shown in the column \textit{Type} where C, D, and M denote the algorithm accepts only continuous, discrete, or mixed type variables. The causal CG generates the next configuration to try to search the configuration space. It is currently implemented as a simple grid search. CL currently employs the causal discovery implementations from the Tetrad (https://github.com/cmu-phil/tetrad) and the Tigramite (https://github.com/jakobrunge/tigramite) projects. In the future, we aim to include more causal discovery implementations.

The causal CE system is responsible for evaluating the configurations and identifying the winner. Unlike AFS, {\em the problem of fitting a causal model is unsupervised}. This means that a standard cross-validation procedure cannot be applied directly. Several methodologies have been devised for causal model selection and/or causal configuration optimization \cite{Raghu2018EvaluationOC, Sedgewick2016LearningMG, Maathuis2009EstimatingHI}. In this work, we employ the OCT (Out-of-sample Causal Tuning) algorithm proposed in \cite{Biza2022OutofSampleTF} shown to outperform other methods and is applicable to mixed data types and models admitting latent confounding variables. We now present the main ideas of OCT for completeness. First, we exploit the fact that the true causal graph induces the correct \vars {Mb} for each variable. Hence, it can lead to optimally predictive models for each node in the graph, given its estimated \vars {Mb}. We define the performance of a causal graph as the average predictive performance of each node given its \vars {Mb}. The true causal graph will exhibit the highest predictive performance. The second idea is to estimate the predictive performance of a configuration using an out-of-sample protocol, e.g., cross-validation: for each fold, a causal configuration produces a causal graph, for each node we fit a predictive model given its corresponding \vars {Mb} and  we estimate the predictive performance on the hold-out fold. Extra edges in the graph, lead to estimated \vars {Mb}  that are super-sets of the true \vars {Mb}  and do not affect predictive performance asymptotically. Hence, causal models with extra edges will have the same predictive performance as the true causal graph. To compensate for that, OCT penalizes causal models for large \vars {Mb}. Specifically, OCT selects as the winning configuration the one with the smallest \vars {Mb} on average, whose predictive performance is statistically indistinguishable from the configuration with the highest predictive performance. Statistical indistinguishability is determined by a permutation-based test. The final idea concerns OCT with mixed data types. To estimate the average predictive performance of a graph, we average performances over all nodes. When these are of different data types, one cannot add different types of metrics, e.g., AUC with mean squared error. To enable OCT with mixed data types, we use a metric defined for all data types, namely the mutual information of the outcome with the predictions of the predictive model. The final causal model is produced by executing the winning causal configuration on all available data.

The output of the CL is (i) the winning causal configuration and (ii) the causal model. 

\begin{table}
  \caption{Causal Discovery Algorithms in CL}
  \label{tab:freq}
  \centering
  \begin{tabular}{c|ccc|ccc}
    \toprule
   & Algorithms & Latent & Temp. & Tests /Scores & Type \\
    \midrule
   \parbox[t]{2mm}{\multirow{7}{*}{\rotatebox[origin=c]{90}{Tetrad}}}
   &PC  variants  &  \xmark  &   \xmark & FisherZ  &C   \\
    &  \cite{Spirtes2000, Ramsey2006AdjacencyFaithfulnessAC, Colombo2014OrderindependentCC} &   &    & BIC \cite{schwarz1978}  &  C \\
      
     &FGES  \cite{Chickering2002OptimalSI, Ramsey2016AMV} &  \xmark  &   \xmark  &  Chi-squared  & D \\
     &FCI variants   & \cmark  &   \xmark  & G-squared  & D \\

    &\cite{Spirtes2000, Raghu2018ComparisonOS, Ramsey2006AdjacencyFaithfulnessAC, Colombo2012LearningHD}  &  & &  BDeu \cite{Heckerman1995LearningBN, Chickering1996LearningEC}& D \\

     &GFCI  \cite{ogarrio16} & \cmark  &   \xmark  &CG \cite{Andrews2017ScoringBN} & M\\
     &SVAR-(G)FCI  \cite{Malinsky2018CausalSL} &\cmark  &   \cmark &  DG \cite{Andrews2019LearningHD} & M  \\
    \midrule

    \parbox[t]{2mm}{\multirow{4}{*}{\rotatebox[origin=c]{90}{Tigramite}}}
    & & & &\\
     &PCMCI(+) \cite{Runge2019DetectingAQ, Runge2020DiscoveringCA} & \xmark & \cmark & ParCor \cite{Runge2019DetectingAQ}& C  \\
     & &   & & G-squared & D  \\
    &LPCMCI \cite{Gerhardus2020HighrecallCD}   &  \cmark & \cmark  & Regression CI & M \\
    
  \bottomrule
\end{tabular}
\label{table:causalconfigs}
\end{table}

\subsection{Causal Reasoning and Visualization}
In this section, we describe AutoCD's functionalities given an estimated causal structure. In particular, we describe (i) the computation of edge confidences on the estimated edges, (ii) the visualization of the causal graph and (iii) the identification of causal paths of interest.

\subsubsection{\textbf{Confidence calculations on causal findings}}

CRV uses bootstrapping to calculate the confidence of causal findings in the optimal estimated causal graph  $\hat G$, such as edges and paths. Specifically, we create bootstrap versions of the data and learn a population of corresponding causal graphs, using the optimal causal configuration. In the case of temporal data, we create bootstrapped samples as in \cite{Debeire2023BootstrapAA}. For each causal feature of interest in $\hat G$, e.g., the presence of a directed edge $\var A \longrightarrow \var B$, we measure the percentage of graphs in the population containing the specific edge. {\em This percentage estimates the probability (confidence) that the given causal configuration would output the specific causal finding; it should not be interpreted as the probability the finding is true irrespectively of the causal discovery algorithm used.} Notice that one bootstrap sample may lead to the finding $\var A \longrightarrow \var B$ and another to $\var A \circ \rightarrow \var B$. The first edge is a special case of the second. Both edges are {\em consistent} with each other since their semantics can both be true in the same PAG. In other words, if the edge $\var A \longrightarrow \var B$ is found 40\% of the times, and $\var A \circ \rightarrow \var B$ is found another 40\% of the time, then we are \%80 confident in the edge $\var A \circ \rightarrow \var B$. In AutoCD we also report the \textbf{edge consistency frequency} for each edge in the winning causal graph, that is the percentage an edge is found is all of its consistent forms.

\subsubsection{\textbf{Graph visualization}}
In case of large datasets, the interpretation of the above results can be quite challenging. Here, we provide a visualization library to include most of the estimated causal information from the AFS and CL steps. We use the Cytoscape,  an open source software platform for visualizing complex networks, especially biological networks \cite{Shannon2003CytoscapeAS}. Currently, our library report basic causal information, such as the group of nodes that are the causal ancestors of the target and the confidence on the estimated edges.  
We note that Cytoscape is a powerful visualization tool that can be used in causal analysis together with other biological studies. 

\subsubsection{\textbf{Causal Queries}}
Despite our visualization techniques, extracting causal information from a very large graph might still be challenging (i.e. to find the causes of a node of interest or the causal paths that connect two nodes). We provide a list of causal queries to facilitate the above tasks.  For a given causal graph  and two nodes A and B, CRV can answer the following queries:
\begin{itemize}
    \item Is there an edge between A and B and what type? 
    \item Is there a directed path from A to B and which?
    \item Is there a potentially directed path\footnote{A potentially directed causal path is a path that could be oriented into a directed path by changing the circles on the path into appropriate tails or arrowheads \cite{Zhang2008CausalRW}. } from A to B and which?
    \item Is there a path of any type connected A and B and which?
\end{itemize}

\section{Case Study: Telecommunication 5G causal network}
\label{sec:problem}

In this work, we try to induce the causal relations among telecommunications time-series measurements from a commercial 5G network. The data and their source cannot be publicly disclosed, unfortunately. The input dataset contains measurements of 143 variables over 144 hourly intervals over 96 5G network cells. In the analysis, we assume the causalities among the variables are independent of the cell and we pool the data from different cells together. There is a specific variable of interest related to the network throughput to focus on the induction of the causal network. 

The causal analysis of this task is quite challenging. First, the data are {\em temporal} and one needs to represent several time-points of the measurements. After preliminary experimentation with the AFS, we decided to explicitly represent three timepoints of the measurements. In other words, preliminary experiments have shown that the process has Markov order 2 (i.e., the current time-point at time \textit{k} is independent of the past of the process, given the measurements of the previous 2 time-points). The number of {\bf variables} measured in the problem is 143 which, when multiplied by 3, results in 429 measured quantities to be analyzed (hereafter, we will be calling {\bf feature} the measurement of a given variable at a specific time-lag). Causally analyzing temporal data requires imposing a stationarity constraint that reflects on the causal relationships discovered. Specifically, if a causal algorithm detects the edge $\var A_{t-1} \rightarrow \var B_{t}$, then the edge $\var A_{t-2} \rightarrow \var B_{t-1}$ should also be included in the graph, as the causal relations are assumed invariant over time. Hence, to causally analyze the data, we need algorithms specifically designed to impose such constraints. A second challenge is the large dimensionality of the data. Most causal discovery algorithms do not scale up to 429 features, represented in this task. Thirdly, the data types are mixed (continuous and discrete). Most causal discovery algorithms work either with one or the other data type. Fourth, the nature of the problem makes the assumption of causal sufficiency (no latent confounders) unrealistic. Hence, we would like to apply more advanced algorithms that admit latent confounders to get more plausible causal findings.

For the telecommunication problem, we utilize AutoCD's modules as follows. We first apply the AFS module to estimate the \vars {Mb} of the target and reduce the dimensionality of the problem. Next, we select the optimal causal graph with the CL module. With the CRV module, we calculate the edge confidence in our estimations, visualize the graph and report the causal paths of interest. 

\subsection{AFS on the 5G dataset}

We apply the AFS module on the time-lagged dataset over three time-lags ($t-2, t-1, t$) treating it as an atemporal (cross-sectional) dataset with 429 features and 96 cells $\times$ 142 time-points $ \approx 13000$ samples. The target variable is the one related to the throughput of the network. Our goal is to find the Markov boundary of the target at the time lag $t$. We split the dataset in $D_{train}$ and $ D_{test}$,  apply AFS on $D_{train}$, and estimate the predictive performance on $D_{test}$. AFS searches over four predictive configurations, consisting of one predictive learning algorithm (RF), two feature selection algorithms (FBED, SES), and two values ($0.01, 0.05$) of their hyper-parameter, namely the significance level of the termination criterion. The estimated Markov boundary, $\vars{Mb}_{est}$, consists of 27 features. The predictive performance with respect to $R^2$ metric on $ D_{test}$ is 0.23.

\begin{figure*}
\begin{minipage}{.6\textwidth}
  \includegraphics[width=\linewidth]{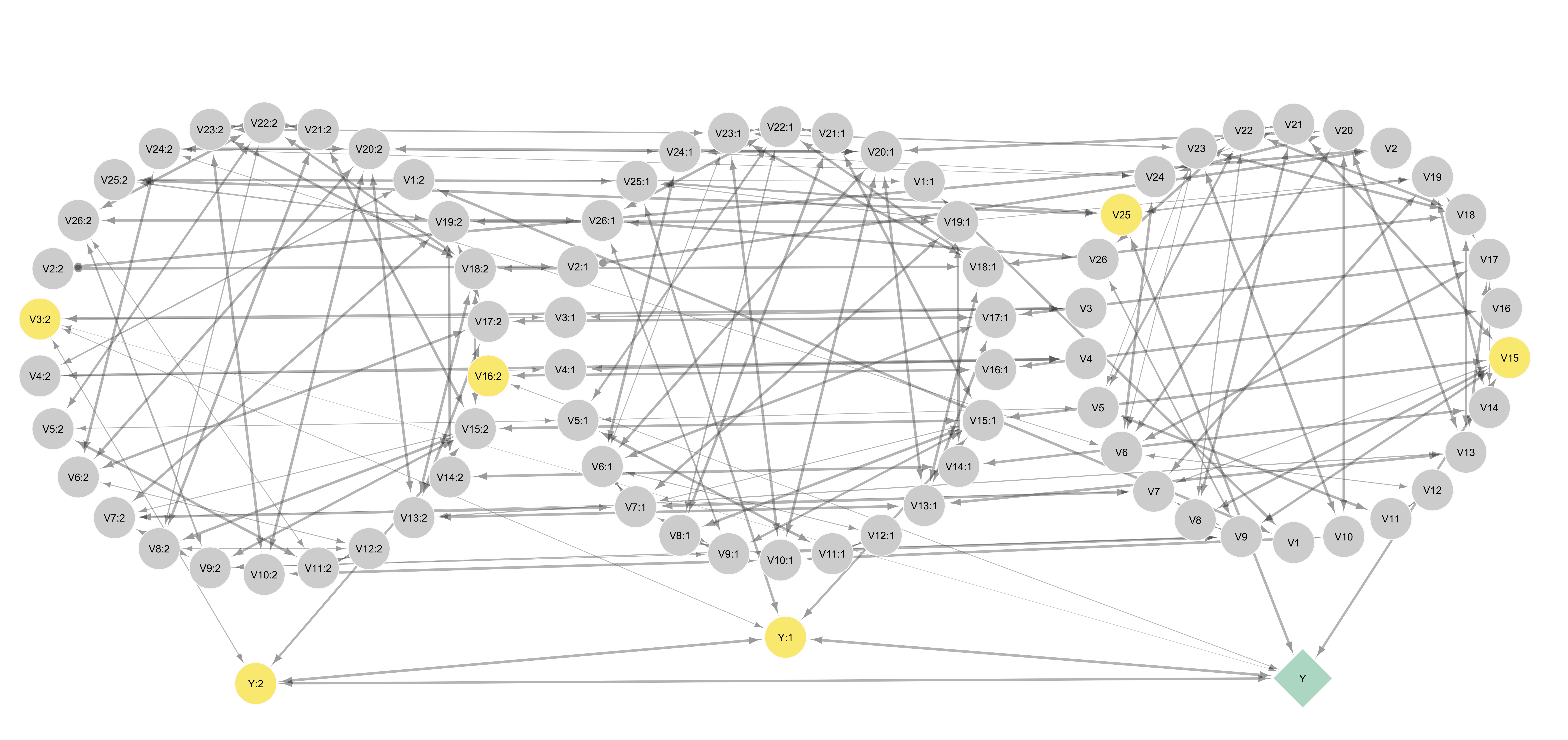}
\end{minipage}%
\hfill
\begin{minipage}{.3\textwidth}
  \centering
  \includegraphics[width=\linewidth]{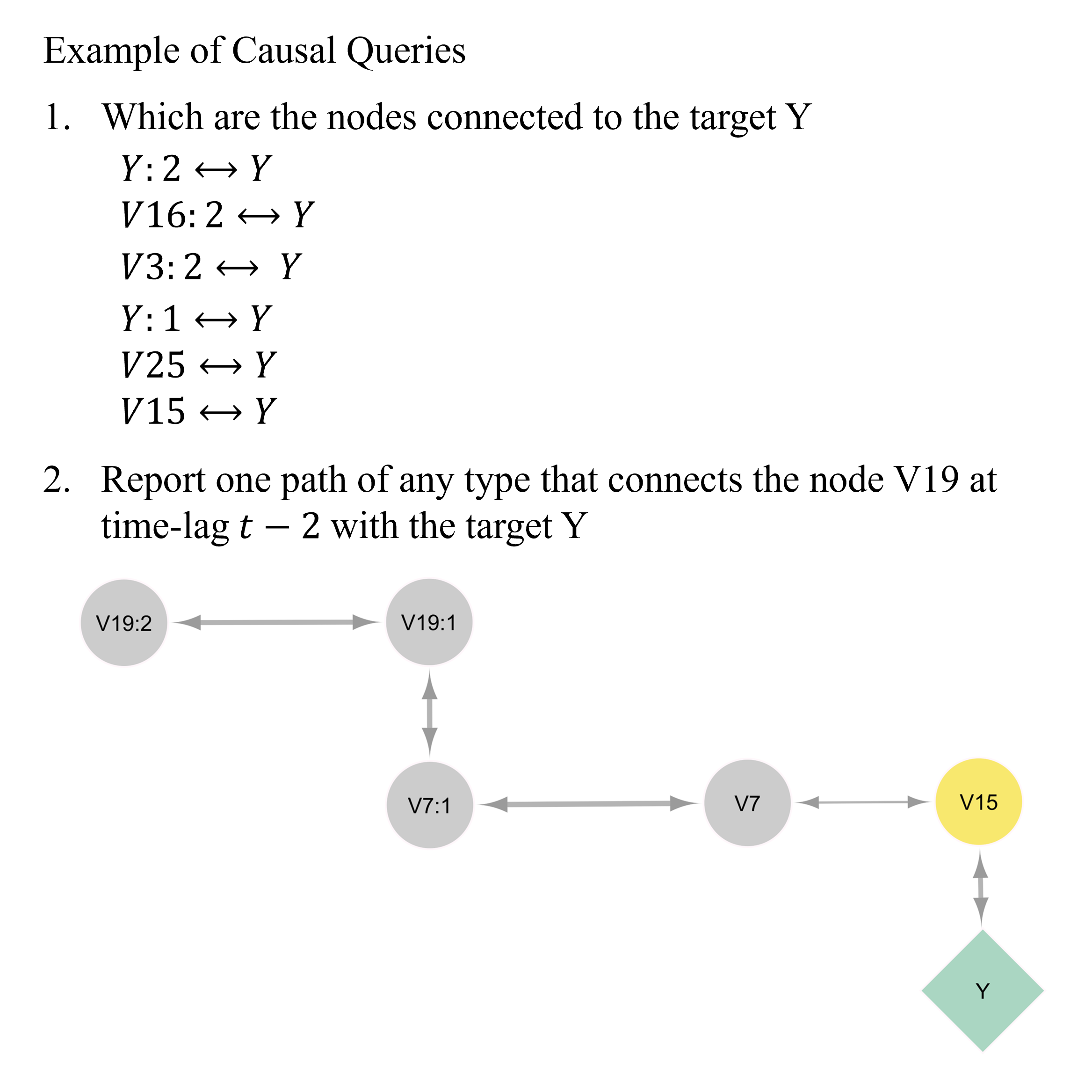}
\end{minipage}

  \caption{The estimated causal structure for the telecommunication problem using the AFS, CL and CRV modules of AutoCD.}
    
  \label{fig:telecomnet}
\end{figure*}

\subsection{CL and CRV on the 5G dataset}

The CL module takes as input the $\vars{Mb}_{est}$ and outputs the optimal time-lagged causal graph. In the causal analysis, we include all 27 features in all 3 time-lags modeled, ending up with 81 features/nodes in the causal graph. We examined a plethora of causal discovery algorithms proposed in the literature, trying to simultaneously address all challenges namely (a) natively treating temporal data and imposing stationarity constraints, (b) scaling to the dimensionality of the problem, (c) accepting mixed data types, and (d) admitting latent confounding variables. Peeking at Table \ref{table:causalconfigs}, the algorithms that satisfy (a), (c), and (d) are SVAR-FCI, SVAR-GFCI, and LPCMCI. Unfortunately, the public implementation of LPCMCI does not scale well to problems with more than 20 features. In addition, we have discovered that the current public implementations of SVAR-FCI and SVAR-GFCI do not produce valid results for problems with more than 10 features (authors have been notified). Hence, we consider FCI and GFCI that address challenges (a), (b), and (c) only, i.e., they do not impose stationarity causal constraints. We employ the OCT tuning method to optimize over these two algorithms and different levels of significance $(0.01, 0.05)$, used as a threshold in their conditional independence test. 

Next, we apply the CRV module to compute edge consistency frequency and visualize the resulting winning causal graph in the form of a PAG. The estimated causal graph, shown in Figure \ref{fig:telecomnet}, contains 163 edges. The width of the edges is proportional to the corresponding edge consistency frequency, to make the most confidence edges stand out. The graph nodes are grouped and laid out according to the time-lag they represent. The target variable is denoted with green and its neighbors with yellow.

Despite dimensionality reduction, the graph is still relatively large for a human expert to inspect and comprehend. CRV allows the user to query the model regarding potentially interesting causal findings. Figure \ref{fig:telecomnet}(right) shows the result of two queries. The first one is what are the nodes connected to the target. As we can see, all such nodes are connected to the target through bi-directed edges, which implies extensive confounding is present. The library also supports the identification of the causal ancestors of the target. However, since all edges are bi-directed, this set is empty. The second query asks for a causal path of any type between the node V19:2 (i.e., variable 19 at time-lag 2) and the target. The library then creates a sub-network with the path of interest, if such exist.

\section{Experimental evaluation on resimulated data}
\label{sec:exp_resim}

To fully evaluate our architecture, we need to apply it to numerous problems, where the causal model that generated the input data is known. This is practically impossible with real data, so we rely on the generation of synthetic data that follow the distribution of known causal graphs. We then evaluate (i) the accuracy of identification of the Markov boundary during the AFS step, (ii) the estimated causal structure and the performance of the tuning method in CL, and (iii) the calculation of the edge confidence compared to the true causal edges.

\begin{figure}
  \centering
  \includegraphics[width=0.6\linewidth]{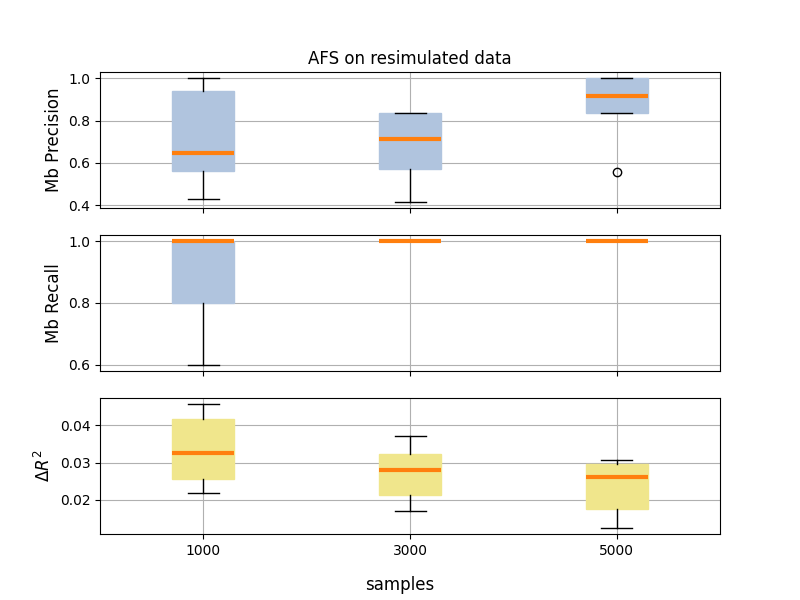}
  \caption{The precision and recall of the estimated Markov boundary and the difference in predictive performance on resimulated data over increasing sample size.}
  \label{fig:resim_MB}
\end{figure}

\subsection{Experimental setup}

We try to use a gold standard causal model that is as close to the unknown graph of the real data. We learn a causal graph $\graph G_{true}$ from the 5G telecommunication data using the FGES algorithm, for three time-lags and a subset of size 41 of the original variables (to reduce computational effort). This is in contrast to several works in the literature, where the causal graph structure is generated stochastically (e.g., each edge is included in the graph with a uniform probability). The functional dependencies between the nodes are assumed linear with additive noise. The coefficients are learned from the original data, and the noise is sampled from the residuals of the linear model, without making parametric assumptions. In other words, we simulate data from a causal structure, coefficients, and noise that are fit on real data to produce a more realistic simulation. This technique is called resimulation \cite{Biza2022OutofSampleTF}. Given the causal model and the resimulation technique above, we create 10 synthetic datasets for each sample size (1000, 3000, 5000), totaling 30 datasets.

\subsection{Evaluation of AFS}
\label{sec:afs_eval}
In the AFS module, we search over four predictive configurations, consisting of one predictive learning algorithm (RF), two feature selection algorithms (FBED, SES) and two significance levels ($0.01, 0.05$). 

AFS returns an estimated Markov boundary $\vars {Mb}_{est}$ of the target at time lag $t$. The corresponding true Markov boundary $\vars {Mb}_{true}$ is determined by $\graph G_{true}$. In Figure \ref{fig:resim_MB}, we plot the precision and recall of Markov Boundary identification. As we increase the sample size, AFS output approaches $\vars {Mb}_{true}$ (recall is one) with some false positives (precision is close to 1). We then compute the difference between the predictive performances (as measured by  $R^2$), called  $\Delta R^2$, between the fitted model $f(\vars {Mb}_{est})$ returned by AFS and the optimal model $f(\vars {Mb}_{true})$ of the gold standard. The larger the difference, the worse the predictive model by AFS. Figure \ref{fig:resim_MB}(bottom) shows that as we increase the sample size, we get closer to the optimal predictive performance.  

\begin{figure}
  \centering
  \includegraphics[width=0.6\linewidth]{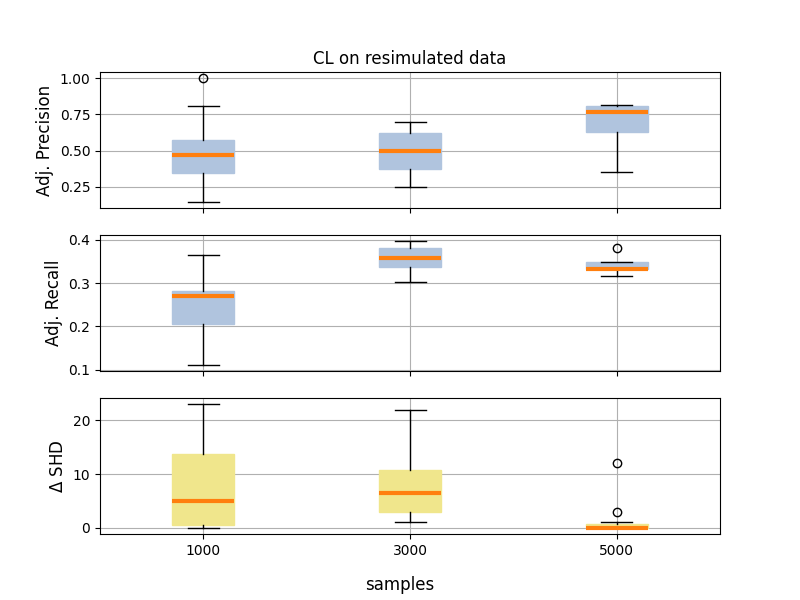}
  \caption{The edge adjacency precision and recall on the selected estimated graph and the difference in SHD on resimulated data over increasing sample size. }
  \label{fig:resim_CL}
\end{figure}

\subsection{Evaluation of CL}

We use the same setup of the CL module as in the case study. The CL module returns the winning time-lagged causal graph $\graph G^{Mb}_{est}$, where the superscript indicates that it is learned only from the variables returned by AFS. We compare this graph with $\graph G^{Mb}_{true}$, which is the marginal of the true graph over the variables of the true Mb (see \cite{richardson2002} for theoretically computing the marginal of a MAG). In Figure \ref{fig:resim_CL}, we show the precision and recall of the adjacencies (i.e., edges ignoring orientation). As we increase the sample size, the adjacency precision increases, however, this is not the case for the adjacency recall. In general, $\graph G^{Mb}_{true}$ is a very dense graph, due to the marginalization procedure that creates numerous confounders.

Regarding the performance of the tuning method, our experiments show that OCT can select the optimal configuration, with respect to the SHD metric. In Figure \ref{fig:resim_CL}, we show the difference in SHD, between the optimal and the selected causal configuration.  SHD counts the
number of steps needed to reach the true PAG from the estimated PAG \cite{Triantafillou2016ScorebasedVC}. As a result, SHD reflects both the adjacency and the orientation errors.

\subsection{Evaluation of CRV}
\begin{figure}
  \centering
  \includegraphics[width=0.6\linewidth]{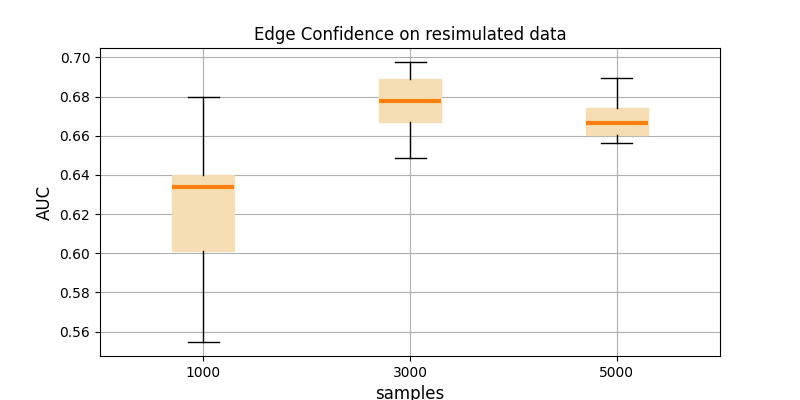}
  \caption{The AUC of the edge consistency frequency on resimulated data over increasing sample size. }
  \label{fig:resim_auc}
\end{figure}

The CRV module takes as input the estimated causal graph and the selected causal configuration. We first evaluate the calculation of the edge consistency frequency using the bootstrapping approach. For each dataset, we apply the selected causal configuration on 100 bootstrapped samples. Using the estimated bootstrapped graphs $\{\graph Gi^{Mb}_{est}\}$, we compute the edge consistency frequency of each edge that appears in $\graph G^{Mb}_{est}$. We use the AUC metric to compare these estimates with the existence of the corresponding edge in $\graph G^{Mb}_{true}$ (an edge in $\graph G^{Mb}_{true}$ has label 1). In Figure \ref{fig:resim_auc}, we show the AUC values over increasing sample size. As we increase the sample size, the AUC is around 0.67. We note that any errors from the previous steps (AFS and CL) are propagated here, since we compare against the true Markov boundary and the corresponding denser graph. 

\section{Experimental evaluation on synthetic data}
\label{sec:exp_synthetic}

\subsection{Experimental setup}
We create synthetic temporal data of 20, 50 and 100 nodes. For the data generation, we assume linear relations and additive normal Gaussian noise. The maximum length of the time lag window is two. In general, we follow the generation process that appear in the numerical experiments in \cite{Gerhardus2020HighrecallCD}: the autocorrelation coefficients are uniformly drawn from $[0.2, 0.9]$ and the coefficients from $[0.1, 0.5]$. We use the Tetrad and the Tigramite projects for the construction of the random DAGs and the data generation, respectively. We create 10 datasets of 2000 samples for each network size and we randomly select a target variable. The average node degree in each time lag is 2 and the maximum node degree is 5. For the following experiments, the true predictive model for these datasets is a linear regression model denoted as $f$. For each repetition and network size, we also simulate hold-out data of 500 samples for the estimation of the predictive performance in the AFS module.

\subsection{Evaluation of AFS}
\begin{figure}
  \centering
  \includegraphics[width=0.6\linewidth]{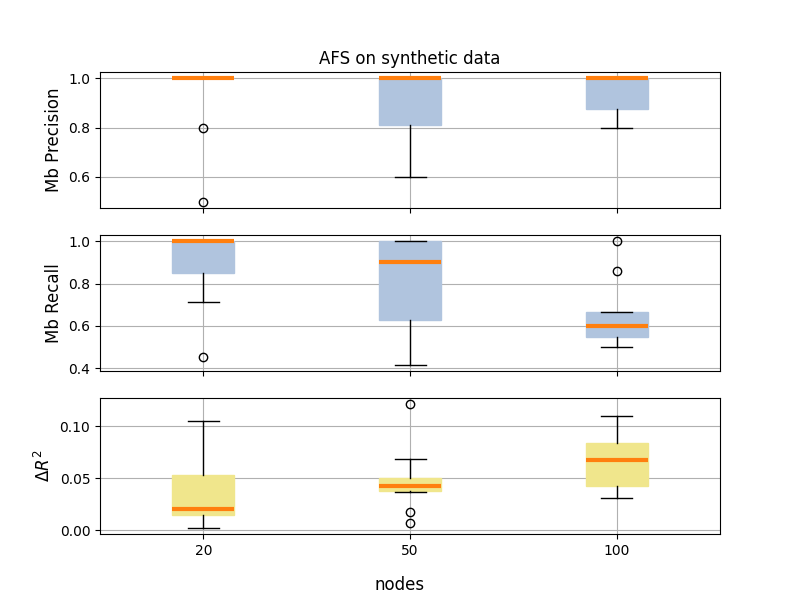}
  \caption{The precision and recall of the estimated Markov boundary and the difference in predictive performance on synthetic data over increasing number of nodes.}
  \label{fig:synthetic_mb}
\end{figure}

In the AFS module, we search over the aforementioned four predictive configurations to estimate the Markov boundary $\vars {Mb}_{est}$ of the target at time lag $t$. We search over three time lags ($t-2$, $t-1$, $t$). In Figure \ref{fig:synthetic_mb}, we plot the results over the increasing number of nodes. The precision is above 0.8 regardless the initial network size. As we increase the network size, the recall is decreasing. This result is expected, since the sample size is remaining constant to 2000 for all network sizes. In addition, we compare the predictive performance on the hold-out samples of the estimated predictive model $\hat{f}(\vars {Mb}_{est})$ with the true model $f(\vars {Mb}_{true})$. In Figure \ref{fig:synthetic_mb}, we report the difference with respect to the $R^2$ metric. The difference is low and increases for larger networks, accordingly.

\subsection{Evaluation of CL}

We apply the OCT tuning method on the datasets that contain only the variables found in $\vars{Mb}_{est}$.  Each variable in $\vars{Mb}_{est}$ appears in each time lag, as in the previous experiments. We now study the causal discovery algorithms that are suitable for time-series data, the SVAR-FCI, SVAR-GFCI and LPCMCI. We include the independence test FisherZ and ParCor with two singinifcance levels $(0.01, 0.05)$. Overall, we apply the OCT tuning method on six causal configurations.
OCT returns the selected time lagged graph $\graph G^{Mb}_{est}$.

\begin{figure}
  \centering
  \includegraphics[width=0.6\linewidth]{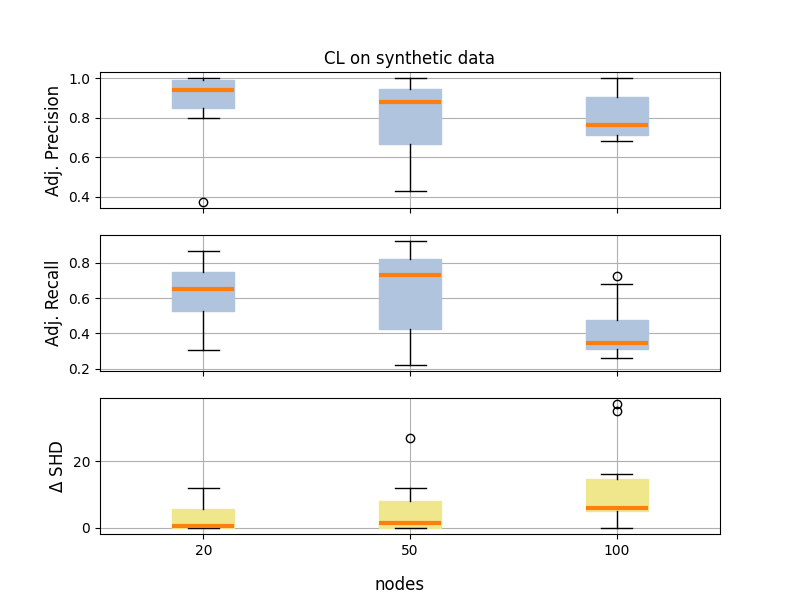}
  \caption{The edge adjacency precision and recall on the selected estimated graph and the difference in SHD on synthetic data over increasing number of nodes.}
  \label{fig:synthetic_cd}
\end{figure}

In Figure \ref{fig:synthetic_cd}, we show the adjacency precision and recall of the edges over the initial network size. The adjacency precision is above 0.8 and slightly descreases on larger networks, following  the corresponding decrease on the Markov boundary precision from the previous AFS step. On the contrary, the adjacency recall is around 0.6 and decreases to 0.4 for the largest network. Similar to the previous experiments,  the $\graph G^{Mb}_{true}$ is a very dense graph, due to the marginalization of many variables. The selected estimated causal graphs miss a number of these edges.  Low recall appears in our experiments, using both groups of causal discovery algorithms (appropriate for cross-sectional or time-series data). However, in the case of time-series causal discovery algorithms, we observe better recall compared to our previous experiments (with the cross-sectional algorithms). This result is in accordance with the work in \cite{Gerhardus2020HighrecallCD}, which shows that FCI and its variants suffer from low recall in autocorrelated temporal data. In Figure \ref{fig:synthetic_cd}, we also show the difference in the SHD metric between the optimal and the selected causal configuration.

\subsection{Evaluation of CRV}
\begin{figure}
  \centering
  \includegraphics[width=0.6\linewidth]{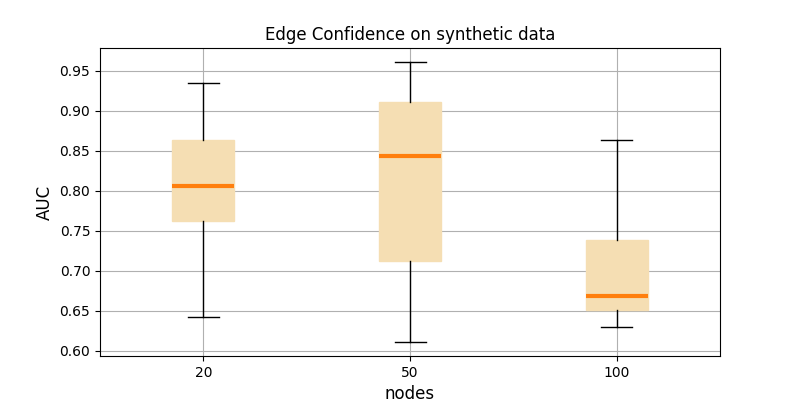}
  \caption{The AUC of the edge consistency frequency on synthetic data over increasing number of nodes. }
  \label{fig:synthetic_auc}
\end{figure}

Given the selected estimated causal structure $\graph G^{Mb}_{est}$ and the corresponding causal configuration, we compute the edge consistency frequency of the estimated edges. We compute the AUC as before, comparing the estimated frequencies with the existence of the corresponding edge in $\graph G^{Mb}_{true}$. In Figure \ref{fig:synthetic_auc}, we show the AUC values over the network sizes. For smaller network sizes, we achieve AUC around 0.8,  which is higher than our previous experiment (with the 41 nodes). This result is in accordance with the corresponding values observed on the adjacency recalls.

\section{Limitations, Discussion, and Future work}
\label{sec:future}

In this work, we propose AutoCD, a causal discovery framework to provide a solution for real-world applications on causal discovery. We propose a pipeline for large-scale temporal data and we use a real-world telecommunication dataset as a case study. Moreover, we extensively evaluate the current methodologies of AutoCD on both resimulated and synthetic data, and we show that they perform well in both cases.  The code of AutoCD is avalaible on our Github webpage \footnote{https://github.com/mensxmachina}, accompanied by the synthetic datasets. 

In the future, we aim to enrich AutoCD modules with several functionalities and more causal discovery algorithms. For all modules, we aim to contain more publicly available packages (e.g. for feature selection and causal discovery). For the AFS module, we aim to (i) search for the optimal data representation and the number of previous time-lags in the predictive configuration space and (ii) use the Bootstrap-Bias Corrected (BBC) \cite{Tsamardinos2018BootstrappingTO} protocol to remove the bias in performance due to trying multiple configurations.
For the CRV module, we aim to include:
(i) an optimization approach for the outcome of interest using the estimated causal graph, (ii) an evaluation procedure for the validity of the causal estimations and (iii) a measure of confidence on the estimated causal paths, similar to the edge consistency frequency.
Finally, an important next step for AutoCD is to construct a web-based platform to facilitate its use by a non-expert user.

\paragraph{Acknowledgments}
We thank Nikolaos Gkorgkolis for his comments on the paper. Funded by the European Union (ERC, AUTOCD, project number 101069394). Views and opinions expressed are however those of the author(s) only and do not necessarily reflect those of the European Union or the European Research Council Executive Agency. Neither the European Union nor the granting authority can be held responsible for them.

\bibliographystyle{plainnat}
\bibliography{references}

\end{document}